\documentclass[10pt,twocolumn,letterpaper]{article}

\usepackage{cvpr}              
\usepackage{multirow}

\usepackage[dvipsnames]{xcolor}

\definecolor{cvprblue}{rgb}{0.21,0.49,0.74}
\usepackage[pagebackref,breaklinks,colorlinks,citecolor=cvprblue]{hyperref}

\def\paperID{1895} 
\def\confName{CVPR}
\def\confYear{2024}

\title{PadChannel: Improving CNN Performance through Explicit Padding Encoding}

\author{Juho Kim\\
Faculty of Applied Science and Engineering\\
University of Toronto, Toronto, ON, Canada\\
{\tt\small juho.kim@mail.utoronto.ca}
}

\begin{document}
\maketitle
\begin{abstract}
In convolutional neural networks (CNNs), padding plays a pivotal role in preserving spatial dimensions throughout the layers. Traditional padding techniques do not explicitly distinguish between the actual image content and the padded regions, potentially causing CNNs to incorrectly interpret the boundary pixels or regions that resemble boundaries. This ambiguity can lead to suboptimal feature extraction. To address this, we propose PadChannel, a novel padding method that encodes padding statuses as an additional input channel, enabling CNNs to easily distinguish genuine pixels from padded ones. By incorporating PadChannel into several prominent CNN architectures, we observed small performance improvements and notable reductions in the variances on the ImageNet-1K image classification task at marginal increases in the computational cost. The source code is available at \url{https://github.com/AussieSeaweed/pad-channel}
\end{abstract}

\section{Introduction}
\label{sec:intro}

\begin{figure}[t]
  \centering
  \includegraphics[width=\linewidth]{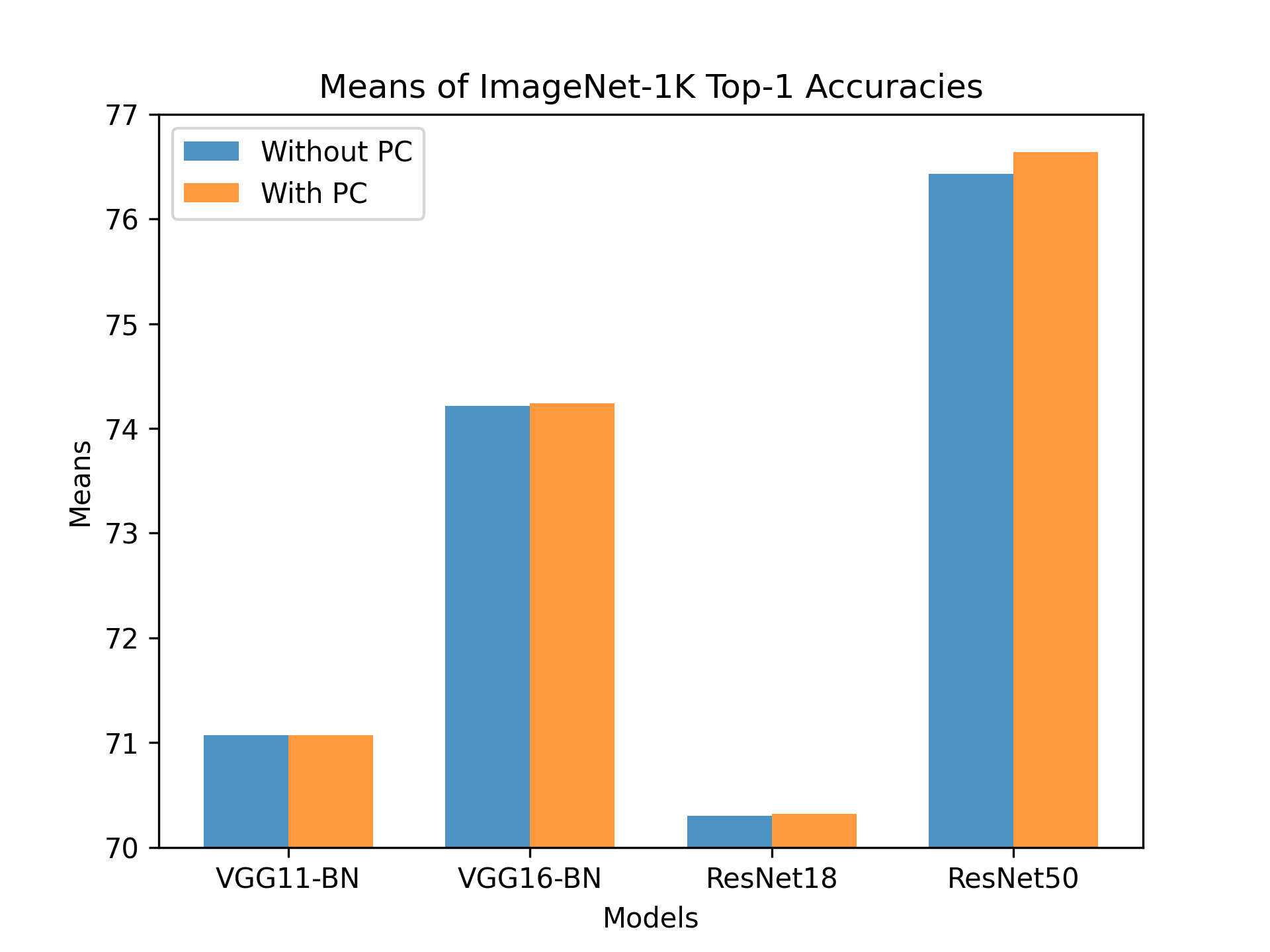}
  \caption{The means of the best single crop top-1 accuracies on the ImageNet-1K validation across each run for each model architecture with and without PadChannel.}
  \label{fig:mean}

  \includegraphics[width=\linewidth]{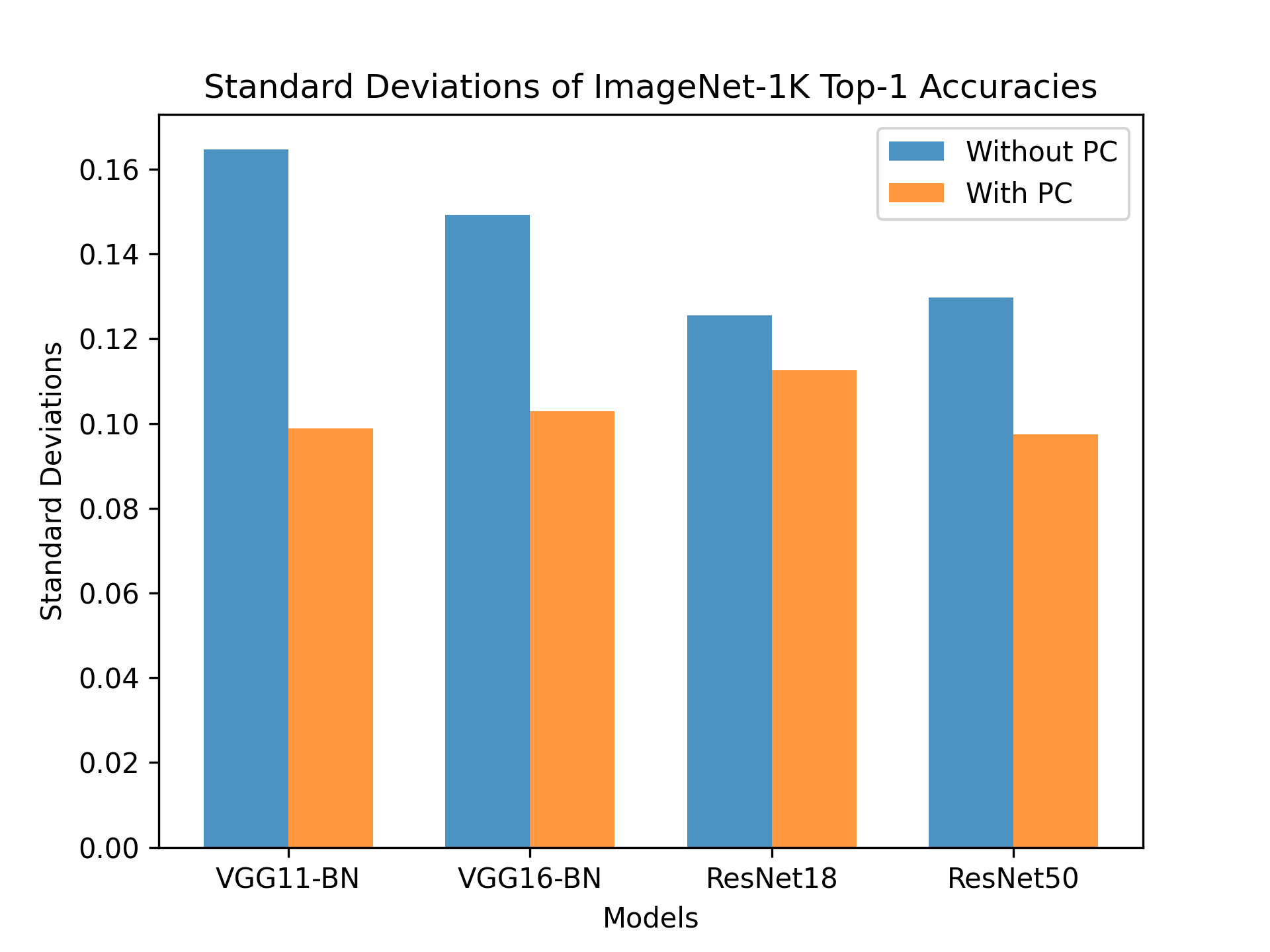}
  \caption{The standard deviations of the best single crop top-1 accuracies on the ImageNet-1K validation across each run for each model architecture with and without PadChannel.}
  \label{fig:stdev}
\end{figure}

In the realm of convolutional neural networks (CNNs), padding plays a pivotal role, ensuring spatial dimensions remain consistent during convolution operations even when part of the kernel lies outside of the original image. A myriad of padding techniques exists, such as zero padding, reflect padding, and replication padding. G. Liu et al. found that, although traditional padding methods often perform at comparable levels, zero padding sometimes emerges as the superior choice \cite{9903574}.

A critical examination of traditional padding techniques like zero padding reveals a significant limitation: such methods do not provide a clear demarcation between the actual image content and the padded regions. As a result, CNNs must discern these boundaries, and while these can be learned, the process is not without ambiguity. This inherent vagueness can potentially hinder optimal feature extraction, emphasizing the need for a more distinct representation \cite{9903574}.

To address this ambiguity when it comes to padding, we developed PadChannel. The PadChannel technique introduces a mask of ones as an extra input channel. When used in combination with zero padding, the values at the extra input channel in the padded areas are zero while the values remain as one for the original image pixels. This clear distinction helps to overcome the limitations of conventional padding methods, significantly improving the ability of CNNs to learn and represent the features accurately with a simple method.

We show that by integrating PadChannel into VGG \cite{simonyan2015deep} and ResNet \cite{7780459} architectures, we achieve performance improvements and reductions in variances on the ImageNet-1K image classification dataset \cite{5206848}, as shown in Figures \ref{fig:mean} and \ref{fig:stdev}, at little to almost no additional computational cost.

In the following sections, we delve into the specifics of PadChannel, its implementation, and a detailed analysis of our experimental results.

\section{Related Works}
\label{sec:rel}

The advent of CNNs has been a breakthrough in the field of computer vision. Central to the functioning of CNNs is the concept of convolution operations, which often uses padding to maintain spatial dimensions across convolutional layers. Commonly adopted padding methods include zero-padding, where boundaries are filled with zeros; reflect padding, where borders mirror adjacent pixels; and replication padding, which extends the closest border pixels \cite{9903574}. These techniques serve the primary purpose of preserving dimensionality but offer no added contextual information for the model, potentially causing CNNs to erroneously interpret the regions near the boundary or regions that resemble image boundaries. This ambiguity can lead to incorrect feature extraction.

Recognizing these limitations, researchers have sought innovative padding techniques to address these challenges. A notable contribution is by G. Liu et al., who introduced partial convolution-based padding. This method involves re-weighing the convolution results near image boundaries. The convolutional weights are calculated using a mask that spans the entire input image \cite{9903574}. Our proposed PadChannel borrows from this idea, yet presents a distinct mechanism by passing it as a separate input channel to the first convolutional layer.

In the domain of CNNs, the concept of using masks as input channels has garnered considerable attention. For example, deepfillv2, proposed by Yu et al. for the image inpainting task, augments the input image with two additional input channels, one to denote holes in the image and another as a user sketch to guide the inpainting process \cite{yu2019free}. The success of DeepFillV2 underscores the potential advantages of providing CNNs with more explicit spatial information. In the context of PadChannel, the extra input channel can help CNNs easily infer which part of the kernel lies outside of the original image thereby giving it a rough positional orientation.

There have been various attempts to directly supply positional information to enhance spatial awareness in CNNs. For instance, R. Liu et al. introduced CoordConv, which encodes the horizontal and vertical positions as distinct input channels. This results in the first CNN layer accepting five input channels: three dedicated to the RGB data and two for the x and y coordinates, respectively \cite{DBLP:conf/nips/LiuLMSFSY18}. Outside of traditional CNNs, Dosovitskiy et al. proposed the vision transformer (ViTs) for various computer vision tasks. ViTs divide the image into fixed-size patches and linearly embed them into vectors. A key feature is their use of positional embeddings, which aids spatial awareness by associating groups of pixels with their relative positions in the image \cite{DBLP:conf/iclr/DosovitskiyB0WZ21}.

These groundbreaking strategies lay the foundation for our novel padding method: PadChannel. This method is designed to mitigate the challenges of traditional padding, offering a more explicit and efficient mechanism for CNNs to interpret image boundaries, at negligible extra computation cost.

\section{Methodology}

\begin{figure*}[t]
  \centering
  \includegraphics[width=\linewidth]{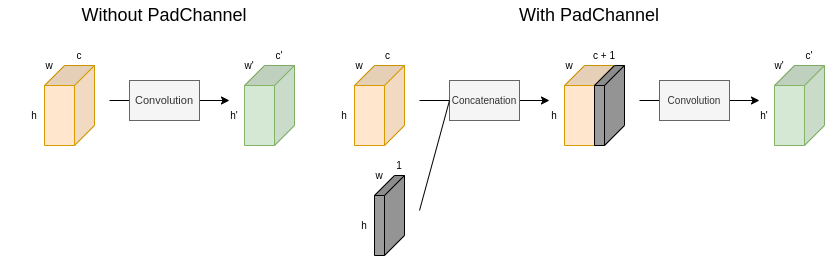}
  \caption{Diagrams of the first convolutional layer with or without PadChannel.}
  \label{fig:pc}
\end{figure*}

\subsection{Model Architectures}
We selected four popular CNN architectures for exploration: VGG11-BN, VGG16-BN \cite{simonyan2015deep}, ResNet18, and ResNet50 \cite{7780459}. These models were chosen not only due to their popularity and widespread use in computer vision but also thanks to their diversity in size and architecture, allowing us to study the general applicability of PadChannel. In addition, the official PyTorch recipes for the aforesaid models are identical, which makes them convenient to train and compare \cite{torchvision2016}.

PadChannel was integrated by modifying the first convolutional layer of each model to accept four input channels instead of the standard three RGB input channels. During the forward pass, a mask of one is concatenated with the input image batch as the new input channel in addition to the RGB values. With the subsequent application of zero padding, the values in the fourth channel corresponding to the padded regions become 0, while regions containing the original pixels retain the value 1. The modified input image transformation at the first CNN layer can be represented as \[X = [X, 1]\] where $X$ is the input image, as visualized in Figure \ref{fig:pc}. With this additional information carried by the fourth input channel, CNNs can more directly understand and interpret boundary pixels.

This clearer demarcation aids the network in focusing on genuine image content and reduces potential pitfalls associated with interpreting padded regions. Note that this makes PadChannel only compatible with zero padding. Other traditional padding techniques such as reflect padding will retain the value $1$ in the padded regions, which defeats the whole purpose of introducing a new fourth channel to clearly mark the regions corresponding to the original image pixels from the padding.

While this technique was only applied on the first layer, this technique can technically be applied to all convolutional layers. The design choice of only modifying the first layer was informed by two reasons. First, adding an extra input channel to every convolutional layer will add a non-trivial computational cost to the overall network. Second, the behavior of PadChannel or encoding of certain positional information can be learned in the subsequent layers. For instance, PadChannel can be achieved in the later layers through the zeroing out of certain weights and the presence of a non-zero bias corresponding to certain output channels. By only applying PadChannel to the first layer, we provide the network with additional useful information while avoiding a significant increase in the computational cost.

To ensure the consistency of our experiments, the modified convolutional layers are initialized with weights using Kaiming initialization \cite{7410480}, as typically done for convolutional layers in PyTorch.

\subsection{Dataset and Data Augmentation}
Our experimentations utilize the ImageNet-1K dataset, a popular choice for image classification tasks, due to its extensive and diverse range of images \cite{5206848}. ImageNet-1K provides over a million labeled images across one thousand different classes, offering a comprehensive and varied set of data crucial for training robust and accurate models. This diversity in the dataset helps in reducing model bias and improves generalizability, making it an ideal choice for developing and benchmarking image classification algorithms.

The choice of ImageNet-1K is also motivated by its widespread use in the machine learning community. Its standardization and consistent annotation process ensure reliability and reproducibility in our experiments.

For data augmentation, we relied on the default transformations used by PyTorch's classification reference scripts \cite{torchvision2016}. These augmentations, which include resizing, random crops, random horizontal flips, and color normalizations during training, and resizing, centered crops, and color normalization during evaluations, are crucial for preventing overfitting and improving model robustness. The choice of these specific augmentations is guided by their proven effectiveness in enhancing the performance of deep learning models on the ImageNet dataset. By simulating various real-world scenarios and introducing variability, these transformations help in training models that are not only accurate but also resilient to variations in new, unseen images.

\subsection{Training Strategy}
Models were trained using the official PyTorch recipes for each respective architecture, guaranteeing standardized training procedures, with the exceptions of the learning rate and the number of epochs. The learning rate was scaled from the default 0.01 to 0.00125 as we trained the models on a single GPU instead of eight GPUs as the recipe was designed for. Then, the learning rate was divided by 10 every 30 epochs. As done by Wu et al. and G. Liu et al., the models were trained for 100 epochs instead of the standard 90 \cite{GroupNorm2018, 9903574}.

To ensure unbiased comparisons, the PadChannel-integrated models and their regular counterparts were trained under identical conditions. To account for variability during training, each network was trained 5 times, as done by G. Liu et al., leading to a total of 40 sets of trained ImageNet-1K model weights \cite{9903574}. In this paper, other traditional padding methods aside from zero padding are not explored, as they have already been well-reviewed by G. Liu et al., who found zero padding to generally be a good choice among the conventional padding methods \cite{9903574}.

\subsection{Hardware and Software Setup}
The models were trained on a single NVIDIA 4090 GPU with Python 3.11.5, PyTorch 2.0.1, and CUDA 11.7.

\subsection{Evaluation Metrics}
After each epoch, models were evaluated on the ImageNet-1K validation set \cite{5206848}. To provide a high-level view of our models' performances, we report the single crop top-1 accuracies of each run by selecting the model checkpoint that exhibits the highest validation set accuracy.

Additionally, to assess the computational efficiency of PadChannel, we report the total number of parameters and multiply-accumulate operations (MACs) for each model, highlighting the small to negligible increases in the computational cost.

In summary, our methodology is designed to provide a rigorous and unbiased assessment of PadChannel's efficacy compared to the performances of baseline models.

\section{Results}

\begin{table*}
  \centering
  \begin{tabular}{@{}l|ccccc|cccc@{}}
    \toprule
	  \multirow{2}{*}{Architecture} & \multicolumn{5}{c|}{Best Top-1} & \multicolumn{2}{c}{\multirow{2}{*}{Mean}} & \multicolumn{2}{c}{\multirow{2}{*}{Stdev}} \\
    & Run 1 & Run 2 & Run 3 & Run 4 & Run 5 & & & & \\
    \midrule
    VGG11-BN & $71.276\%$ & $71.154\%$ & $70.894\%$ & $70.910\%$ & $71.120\%$ & $71.071\%$ & -- & $0.165\%$ & -- \\
    VGG11-BN-PC & $70.940\%$ & $71.030\%$ & $71.058\%$ & $71.204\%$ & $71.120\%$ & $71.070\%$ & $-0.000\%$ & $0.099\%$ & $-0.066\%$ \\
    \midrule
    VGG16-BN & $74.070\%$ & $74.072\%$ & $74.308\%$ & $74.226\%$ & $74.412\%$ & $74.218\%$ & -- & $0.149\%$ & -- \\
    VGG16-BN-PC & $74.270\%$ & $74.230\%$ & $74.122\%$ & $74.396\%$ & $74.184\%$ & $74.240\%$ & $+0.023\%$ & $0.103\%$ & $-0.046\%$ \\
    \midrule
    ResNet18 & $70.152\%$ & $70.496\%$ & $70.314\%$ & $70.250\%$ & $70.292\%$ & $70.301\%$ & -- & $0.126\%$ & -- \\
    ResNet18-PC & $70.200\%$ & $70.292\%$ & $70.316\%$ & $70.506\%$ & $70.290\%$ & $70.321\%$ & $+0.020\%$ & $0.113\%$ & $-0.013\%$ \\
    \midrule
    ResNet50 & $76.286\%$ & $76.436\%$ & $76.372\%$ & $76.638\%$ & $76.428\%$ & $76.432\%$ & -- & $0.130\%$ & -- \\
    ResNet50-PC & $76.582\%$ & $76.702\%$ & $76.780\%$ & $76.592\%$ & $76.546\%$ & $76.640\%$ & $+0.208\%$ & $0.097\%$ & $-0.032\%$ \\
    \bottomrule
  \end{tabular}
  \caption{The best single crop top-1 accuracies on the ImageNet-1K validation set for each run. The checkpoint that yields the highest such value is selected for each run. The model names with the suffix ``-PC'' denote that the PadChannel was applied to the corresponding architecture.}
  \label{tab:top1}
\end{table*}

The single crop best top-1 validation set accuracies for each run are shown in Table \ref{tab:top1}. We observed that models incorporating PadChannel generally achieved higher or at least similar mean validation set accuracy compared to their counterparts without PadChannel. This trend is observable across different models, though the extent of improvement varies. Specifically, for VGG11-BN, there was an insignificant change in accuracy ($-0.000\%$), with a high p-value of $0.5018$, indicating no statistical significance in the difference observed. Similarly, for VGG16-BN and ResNet18, the increases in accuracy were marginal ($+0.023\%$ and $+0.020\%$, respectively) and the p-values ($0.3928$ for VGG16-BN and $0.3988$ for ResNet18) suggest these differences are not statistically significant.

In contrast, for ResNet50, we observed a more notable improvement in accuracy ($+0.208\%$), and the corresponding p-value of $0.0104$ indicates this improvement is indeed statistically significant at the $5\%$ level. This finding suggests that the implementation of PadChannel in the ResNet50 architecture led to a meaningful enhancement in model performance that is unlikely to be due to chance.

These results are encouraging, as they imply that PadChannel does not detrimentally impact model performance. In fact, in certain cases, it appears to provide additional context for the models, aiding in the interpretation of images. Notably, the greater success of PadChannel in ResNet architectures, as evidenced by the statistical significance in the case of ResNet50, hints at the possibility that the residual connections in these networks might be more effective in utilizing the information provided by the fourth channel in subsequent convolutional layers. The varying degrees of improvement and their statistical significance across different architectures emphasize the need for further investigation into how different network designs might interact with additional channels like PadChannel.

In addition, consistent drops in the standard deviations of the validation set accuracies are observed across all network architectures: VGG11-BN ($-0.066\%$); VGG16-BN ($-0.046\%$); ResNet18 ($-0.013\%$); ResNet50 ($-0.032\%$). The presence of PadChannel seems to help stabilize the training process. This observation suggests that the process of learning to discern the boundary regions has a high variability and providing explicit information regarding the padding statuses can facilitate a reduction in this variance.

\begin{table}
  \centering
  \begin{tabular}{@{}l|cc|cc@{}}
    \toprule
    \multirow{2}{*}{Architecture} & \multicolumn{2}{c|}{\# Params} & \multirow{2}{*}{Diff} & \multirow{2}{*}{\% Diff} \\
    & w/o PC & w/ PC & & \\
    \midrule
    VGG11-BN & $132.9$M & $132.9$M & $+576$ & $+0.0004\%$ \\
    \midrule                      
    VGG16-BN & $138.4$M & $138.4$M & $+576$ & $+0.0004\%$ \\
    \midrule
    ResNet18 & $11.7$M & $11.7$M & $+3136$ & $+0.027\%$ \\
    \midrule                    
    ResNet50 & $25.6$M & $25.6$M & $+3136$ & $+0.012\%$ \\
    \bottomrule
  \end{tabular}
  \caption{The number of parameters for each tested architecture with and without PadChannel.}
  \label{tab:nparams}
\end{table}

\begin{table}
  \centering
  \begin{tabular}{@{}l|cc|cc@{}}
    \toprule
    \multirow{2}{*}{Architecture} & \multicolumn{2}{c|}{GMACs} & \multirow{2}{*}{Diff} & \multirow{2}{*}{\% Diff} \\
    & w/o PC & w/ PC & & \\
    \midrule
    VGG11-BN & $7.66$ & $7.69$ & $+0.03$ & $+0.377\%$ \\
    \midrule
    VGG16-BN & $15.55$ & $15.58$ & $+0.03$ & $+0.186\%$ \\
    \midrule
    ResNet18 & $1.83$ & $1.86$ & $+0.04$ & $+2.155\%$ \\
    \midrule
    ResNet50 & $4.13$ & $4.17$ & $+0.04$ & $+0.952\%$ \\
    \bottomrule
  \end{tabular}
  \caption{The Giga Multiply-Accumulate operations (GMACs) for each tested architecture with and without PadChannel.}
  \label{tab:gmacs}
\end{table}

The numbers of parameters in the models with and without PadChannel are shown in Table \ref{tab:nparams}. The percentage increases in the number of parameters due to PadChannel compared to the baseline models are in the hundredths for ResNet models and ten-thousandths for VGG models. The GMACs for each model with and without PadChannel are shown in Table \ref{tab:gmacs}. The table shows marginal increases in the computational cost across all models ranging from $+0.186\%$ for VGG16-BN to $+2.155\%$ for ResNet18.

\begin{figure*}[t]
  \centering
  \includegraphics[width=\linewidth]{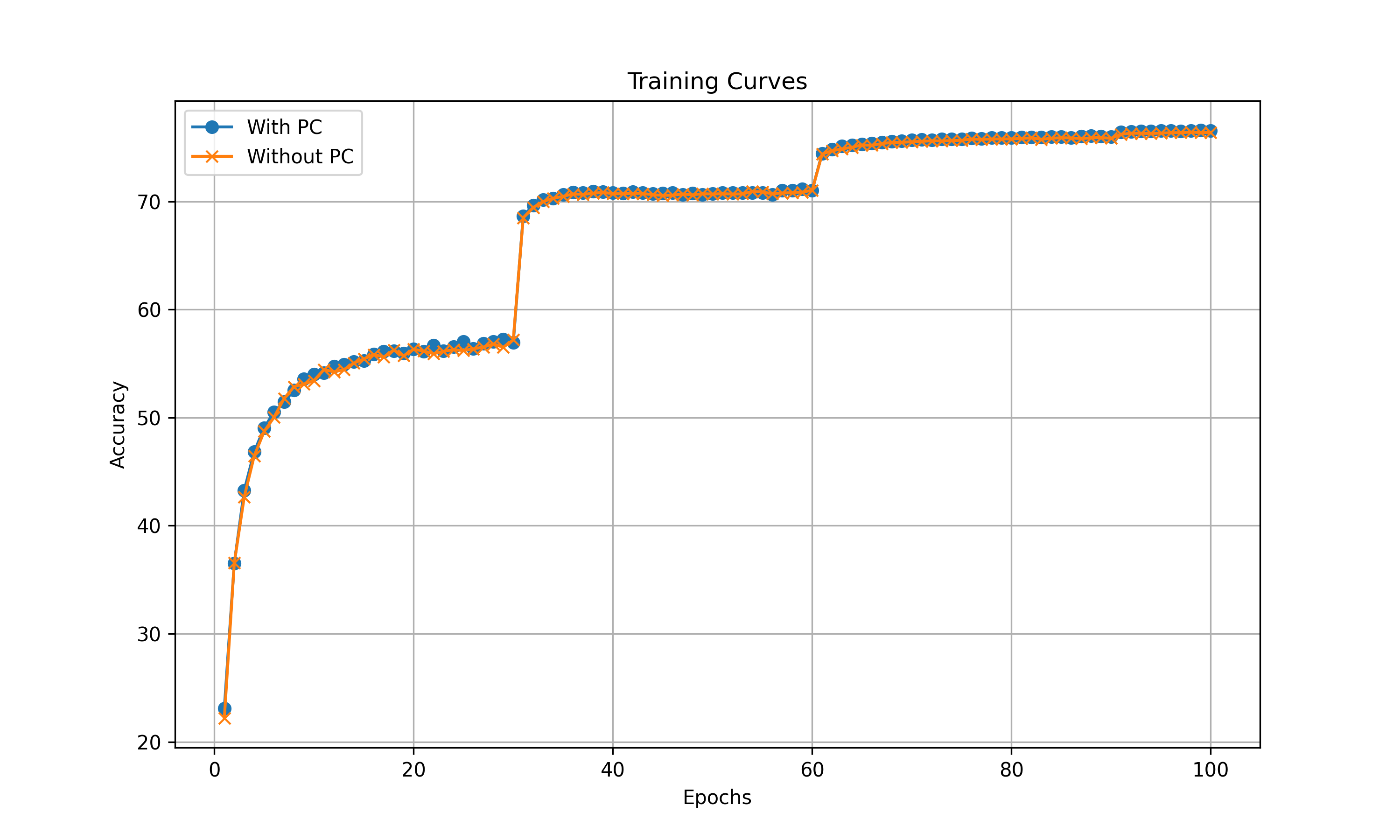}
  \includegraphics[width=0.49\linewidth]{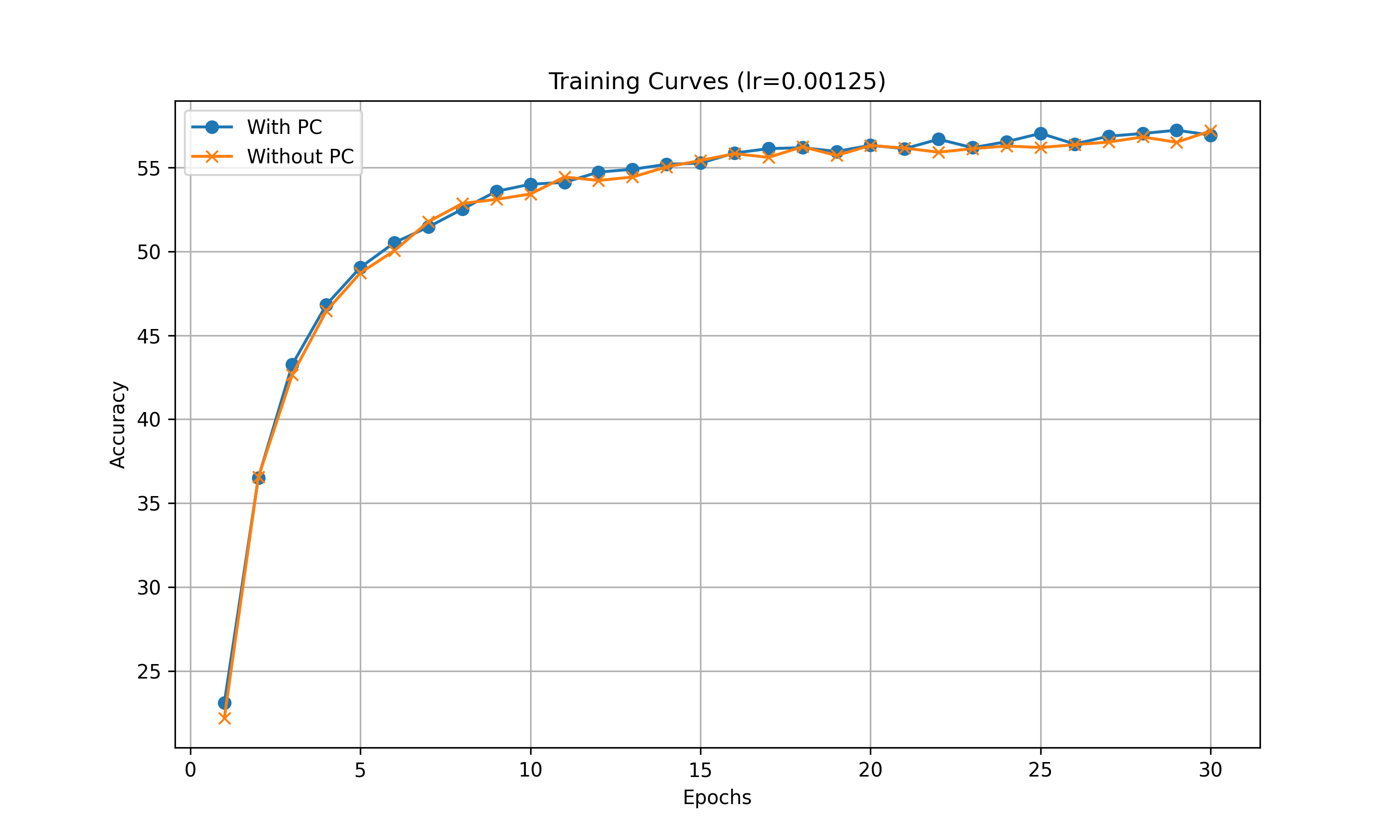}
  \includegraphics[width=0.49\linewidth]{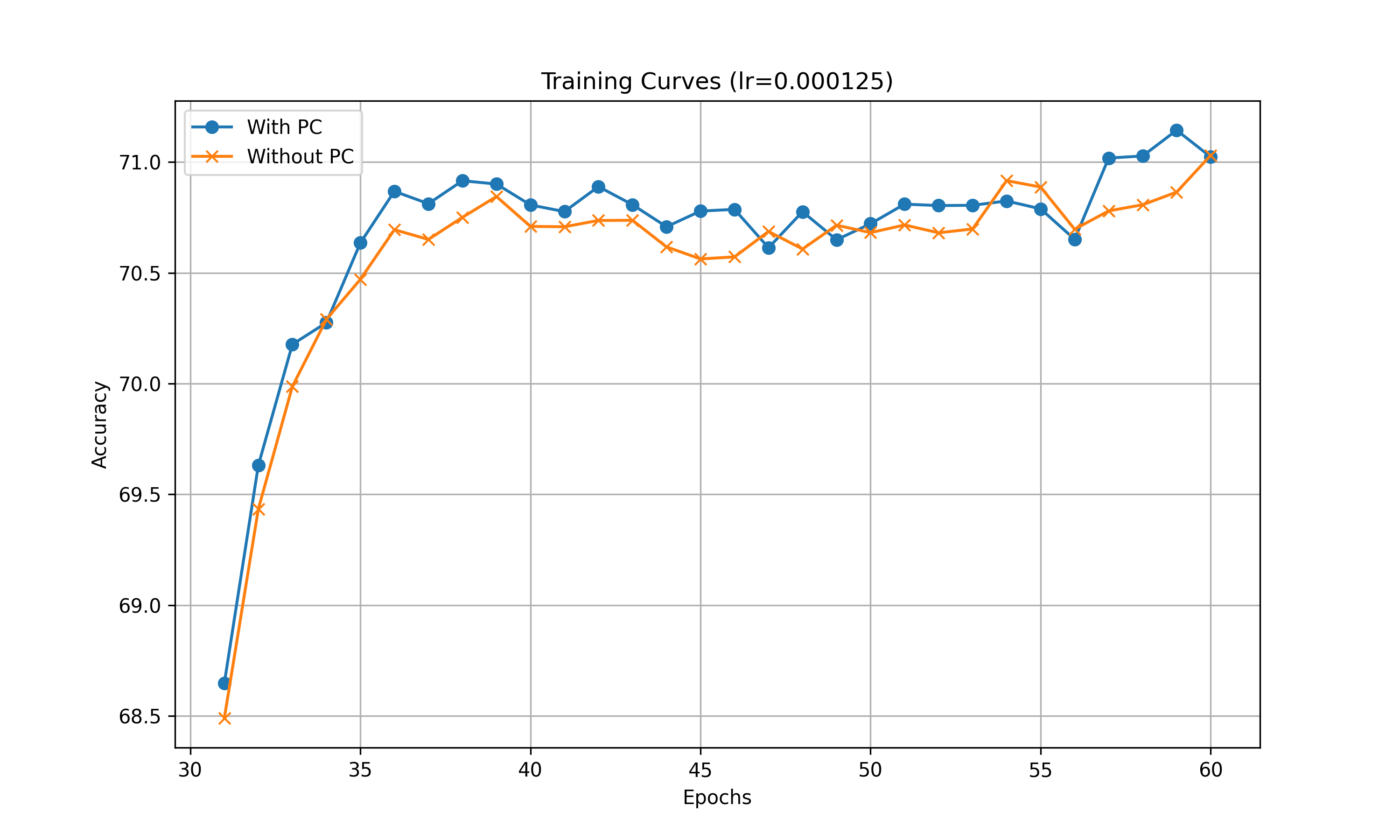}
  \includegraphics[width=0.49\linewidth]{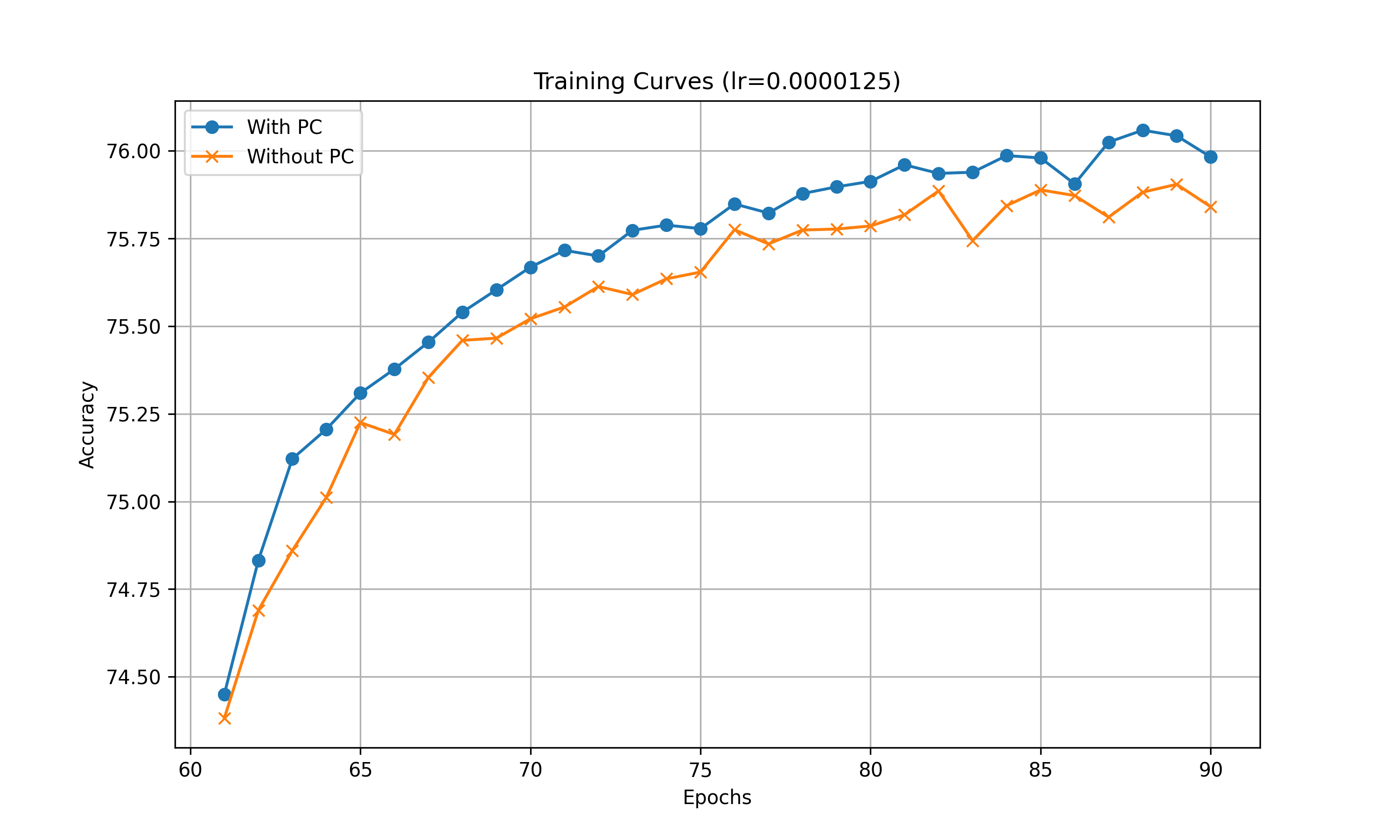}
  \includegraphics[width=0.49\linewidth]{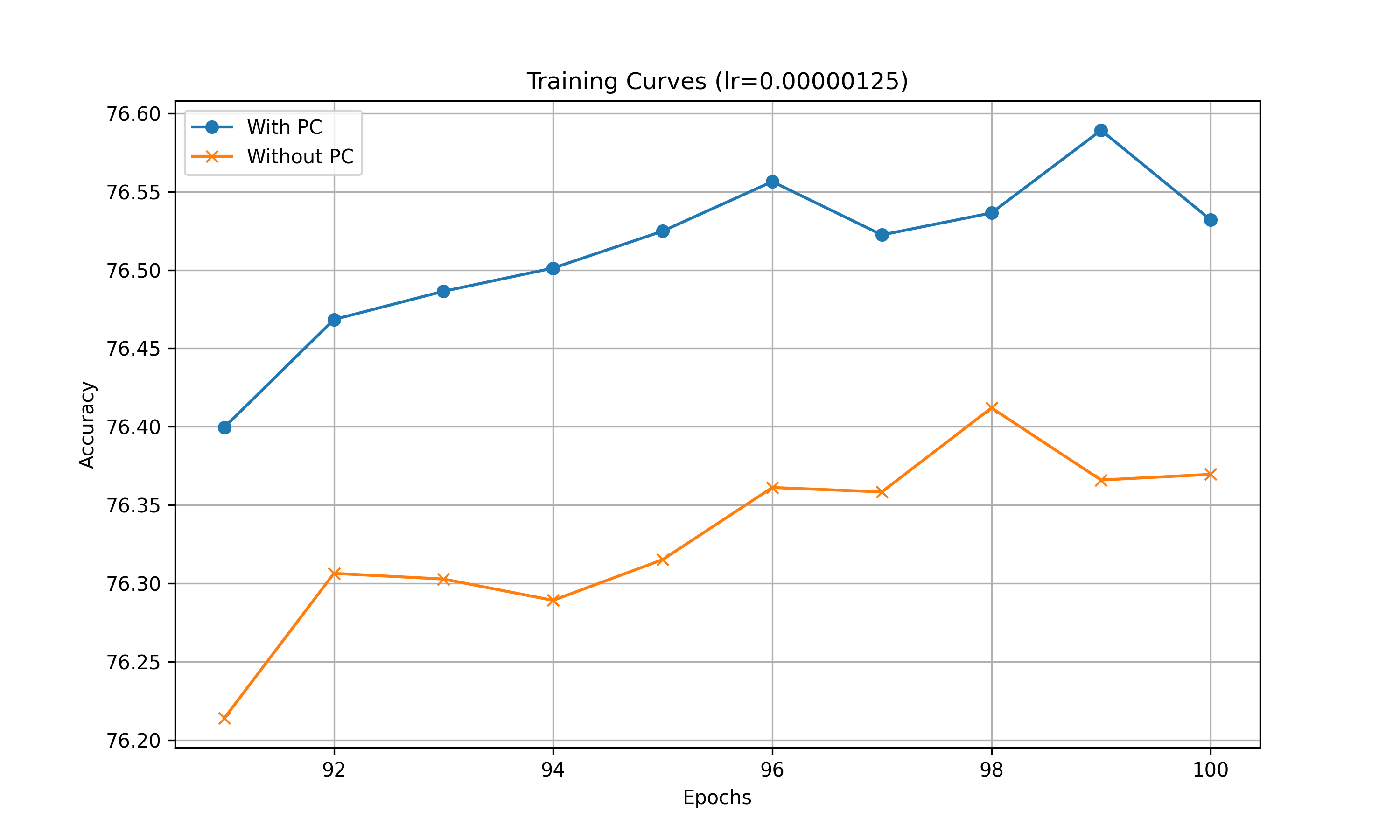}
  \caption{The training curves for the ResNet50 models with and without PadChannel.}
  \label{fig:curves}
\end{figure*}

Figure \ref{fig:curves} illustrates the training curves representing the validation set accuracies at each epoch, averaged across five runs, for the ResNet50 model architecture with and without PadChannel. These curves indicate that models equipped with PadChannel tend to reach specific accuracy benchmarks more quickly than their counterparts without PadChannel. For instance, the curve marked ``with PC'' achieves an accuracy of $75.75\%$ at Epoch $73$, while the ``without PC'' curve reaches the same accuracy level three epochs later, at Epoch $76$. This pattern demonstrates a faster convergence rate for models incorporating PadChannel. The difference may be attributed to the additional fourth channel, which potentially reduces the burden of learning to distinguish the boundary regions. Consequently, the model can more efficiently focus on interpreting the input features that are less influenced by padding.

\section{Conclusion}

This paper presented a novel padding technique: PadChannel. Through our extensive evaluations, we demonstrated that this method not only increases the accuracies of CNN architectures and reduces their variance in the image classification task but also fosters faster convergence compared to the standard models. The efficacy of PadChannel suggests that providing explicit boundary information aids in the network's feature discrimination ability, a hypothesis that invites further theoretical investigation. Due to its flexibility, PadChannel can easily be applied to other CNN architectures from many domains not explored in this paper, including semantic segmentation, object detection, and more by augmenting the corresponding architecture's first convolutional layer to incorporate PadChannel. Exploring the theoretical foundations of PadChannel's effectiveness, alongside its broader impact and applicability, presents promising directions for future research.

\section{Acknowledgment}

We acknowledge the use of ChatGPT, based on GPT-4, an AI language model, in order to enhance the readability of the paper by correcting grammar, polishing language, and improving clarity throughout all sections \cite{openai2023gpt4}.

{
    \small
    \bibliographystyle{ieeenat_fullname}
    \bibliography{main}

\begin{thebibliography}{11}
\providecommand{\natexlab}[1]{#1}
\providecommand{\url}[1]{\texttt{#1}}
\expandafter\ifx\csname urlstyle\endcsname\relax
  \providecommand{\doi}[1]{doi: #1}\else
  \providecommand{\doi}{doi: \begingroup \urlstyle{rm}\Url}\fi

\bibitem[Deng et~al.(2009)Deng, Dong, Socher, Li, Li, and Fei-Fei]{5206848}
Jia Deng, Wei Dong, Richard Socher, Li-Jia Li, Kai Li, and Li Fei-Fei.
\newblock Imagenet: A large-scale hierarchical image database.
\newblock In \emph{2009 IEEE Conference on Computer Vision and Pattern
  Recognition}, pages 248--255, 2009.

\bibitem[Dosovitskiy et~al.(2021)Dosovitskiy, Beyer, Kolesnikov, Weissenborn,
  Zhai, Unterthiner, Dehghani, Minderer, Heigold, Gelly, Uszkoreit, and
  Houlsby]{DBLP:conf/iclr/DosovitskiyB0WZ21}
Alexey Dosovitskiy, Lucas Beyer, Alexander Kolesnikov, Dirk Weissenborn,
  Xiaohua Zhai, Thomas Unterthiner, Mostafa Dehghani, Matthias Minderer, Georg
  Heigold, Sylvain Gelly, Jakob Uszkoreit, and Neil Houlsby.
\newblock An image is worth 16x16 words: Transformers for image recognition at
  scale.
\newblock In \emph{9th International Conference on Learning Representations,
  {ICLR} 2021, Virtual Event, Austria, May 3-7, 2021}. OpenReview.net, 2021.

\bibitem[He et~al.(2015)He, Zhang, Ren, and Sun]{7410480}
Kaiming He, Xiangyu Zhang, Shaoqing Ren, and Jian Sun.
\newblock Delving deep into rectifiers: Surpassing human-level performance on
  imagenet classification.
\newblock In \emph{2015 IEEE International Conference on Computer Vision
  (ICCV)}, pages 1026--1034, 2015.

\bibitem[He et~al.(2016)He, Zhang, Ren, and Sun]{7780459}
Kaiming He, Xiangyu Zhang, Shaoqing Ren, and Jian Sun.
\newblock Deep residual learning for image recognition.
\newblock In \emph{2016 IEEE Conference on Computer Vision and Pattern
  Recognition (CVPR)}, pages 770--778, 2016.

\bibitem[Liu et~al.(2023)Liu, Dundar, Shih, Wang, Reda, Sapra, Yu, Yang, Tao,
  and Catanzaro]{9903574}
Guilin Liu, Aysegul Dundar, Kevin~J. Shih, Ting-Chun Wang, Fitsum~A. Reda,
  Karan Sapra, Zhiding Yu, Xiaodong Yang, Andrew Tao, and Bryan Catanzaro.
\newblock Partial convolution for padding, inpainting, and image synthesis.
\newblock \emph{IEEE Transactions on Pattern Analysis and Machine
  Intelligence}, 45\penalty0 (5):\penalty0 6096--6110, 2023.

\bibitem[Liu et~al.(2018)Liu, Lehman, Molino, Such, Frank, Sergeev, and
  Yosinski]{DBLP:conf/nips/LiuLMSFSY18}
Rosanne Liu, Joel Lehman, Piero Molino, Felipe~Petroski Such, Eric Frank, Alex
  Sergeev, and Jason Yosinski.
\newblock An intriguing failing of convolutional neural networks and the
  coordconv solution.
\newblock In \emph{Advances in Neural Information Processing Systems 31: Annual
  Conference on Neural Information Processing Systems 2018, NeurIPS 2018,
  December 3-8, 2018, Montr{\'{e}}al, Canada}, pages 9628--9639, 2018.

\bibitem[maintainers and contributors(2016)]{torchvision2016}
TorchVision maintainers and contributors.
\newblock Torchvision: Pytorch's computer vision library.
\newblock \url{https://github.com/pytorch/vision}, 2016.

\bibitem[OpenAI(2023)]{openai2023gpt4}
OpenAI.
\newblock Gpt-4 technical report, 2023.

\bibitem[Simonyan and Zisserman(2015)]{simonyan2015deep}
Karen Simonyan and Andrew Zisserman.
\newblock Very deep convolutional networks for large-scale image recognition,
  2015.

\bibitem[Wu and He(2018)]{GroupNorm2018}
Yuxin Wu and Kaiming He.
\newblock Group normalization.
\newblock \emph{arXiv:1803.08494}, 2018.

\bibitem[Yu et~al.(2019)Yu, Lin, Yang, Shen, Lu, and Huang]{yu2019free}
Jiahui Yu, Zhe Lin, Jimei Yang, Xiaohui Shen, Xin Lu, and Thomas~S Huang.
\newblock Free-form image inpainting with gated convolution.
\newblock In \emph{Proceedings of the IEEE International Conference on Computer
  Vision}, pages 4471--4480, 2019.

\end{thebibliography}
}
\end{document}